\documentclass{article}

\usepackage[preprint]{neurips_2026}

\usepackage[utf8]{inputenc} % allow utf-8 input
\usepackage[T1]{fontenc}    % use 8-bit T1 fonts
\usepackage{hyperref}       % hyperlinks
\usepackage{url}            % simple URL typesetting
\usepackage{booktabs}       % professional-quality tables
\usepackage{amsfonts}       % blackboard math symbols
\usepackage{nicefrac}       % compact symbols for 1/2, etc.
\usepackage{microtype}      % microtypography
\usepackage{xcolor}         % colors
\usepackage{graphicx}
\title{LLM-based Conversational AI Knowledge Assistant for MyBuddy Humanoid Robot}

\author{%
  Hanxiao Chen\thanks{Hanxiao Chen is currently the first-year PhD Student in The University of Tokyo, her research mainly focuses on AI, Robotics and Human-Robot Interaction.} \\
  Department of Electric Engineering and Information Systems\\
  The University of Tokyo \\
  Tokyo, Japan 113-8654 \\
  \texttt{hanxiaochen@g.ecc.u-tokyo.ac.jp} \\
}

\begin{document}

\maketitle

\begin{abstract}
Humanoid robots are increasingly being popular and developed for human-centered applications, yet their ability to provide intelligent conversations and natural interactive knowledge assistance remains constrained by traditional rule-based dialogue systems, pre-defined responses and limited knowledge repositories. Large language models (LLMs) have emerged as a powerful foundation for enabling natural, adaptive, and context-aware Human-Robot Interaction (HRI), which provides a significant opportunity to address such limitations by enabling robots to understand natural speech language, reason over complicated queries, maintain high-quality conversational context, and generate knowledge-rich responses. In this work, we originally present and implement an LLM-based versatile Conversational AI Knowledge Assistant for the Raspberry-Pi-powered 13-Axis MyBuddy humanoid robot, which integrates LLM-driven language understanding and AI reasoning with real-time speech recognition, knowledge retrieval via extensible access of internet engines (e.g., Wikipedia, arXiv), flexible dialogue management, and natural speech synthesis to enable much more intelligent multi-turn continuous conversations and advanced emotional-support Human-Robot Interaction.

\end{abstract}

\section{Introduction}
With the rapid advancement of intelligent robotic systems, Human-Robot Interaction (HRI) has increasingly emerged as a significant interdisciplinary research domain, focusing on how humans and robots communicate, interact and collaborate in complicated real-world environments. Driven by Artificial Intelligence, most recent work has contributed to HRI with a broad spectrum of interaction modalities including natural speech dialogue systems [1], haptic tactile touch frameworks [2], text UI interfaces [3] and visual gesture interactions [4], even significantly stimulating the new modern development of multi-modal AI paradigm for robotic manipulation [5] [6]. Moreover, humanoid robots have attracted increasing attention as a promising platform for human-centered AI applications, including education, healthcare, industry, and social interaction. Unlike conventional service robots that primarily execute predefined tasks, humanoid robots are expected to communicate with people in a natural and adaptive manner while providing useful information and assistance. However, traditional HRI systems largely rely on the rule-based conversation systems, predefined response templates, or task-specific knowledge repositories, which exactly limit the robot’s ability to understand diverse natural-language expressions, maintain contextual information across multiple conversational turns, and generate flexible responses to open-ended user queries.

Most recent advances in large language models (LLMs) have introduced new opportunities to solve the proposed limitations and develop more natural intelligent conversational HRI systems since LLMs can provide strong capabilities in natural-language understanding, contextual reasoning, knowledge retrieval, and response generation, enabling robots to move beyond rigid command-response paradigms toward adaptive and context-aware interactions. In this work, we present a versatile LLM-based Conversational AI Knowledge Assistant for the Raspberry-Pi-powered 13-axis MyBuddy humanoid robot. The proposed system integrates LLM-driven natural-language understanding and reasoning with real-time speech recognition, knowledge retrieval, and natural speech synthesis, enabling the robot to interact with users through continuous multi-turn speech conversations. By combining internal and external internet sources (e.g., Wikipedia, arXiv), our system can respond to any open-ended questions while maintaining conversational context and providing more informative and adaptive emotional-support knowledge assistance than conventional rule-based approaches, even can be flexibly transferred to other physical robot platforms.

\section{Method}

\begin{figure}
  \centering
  \includegraphics[width=\textwidth]{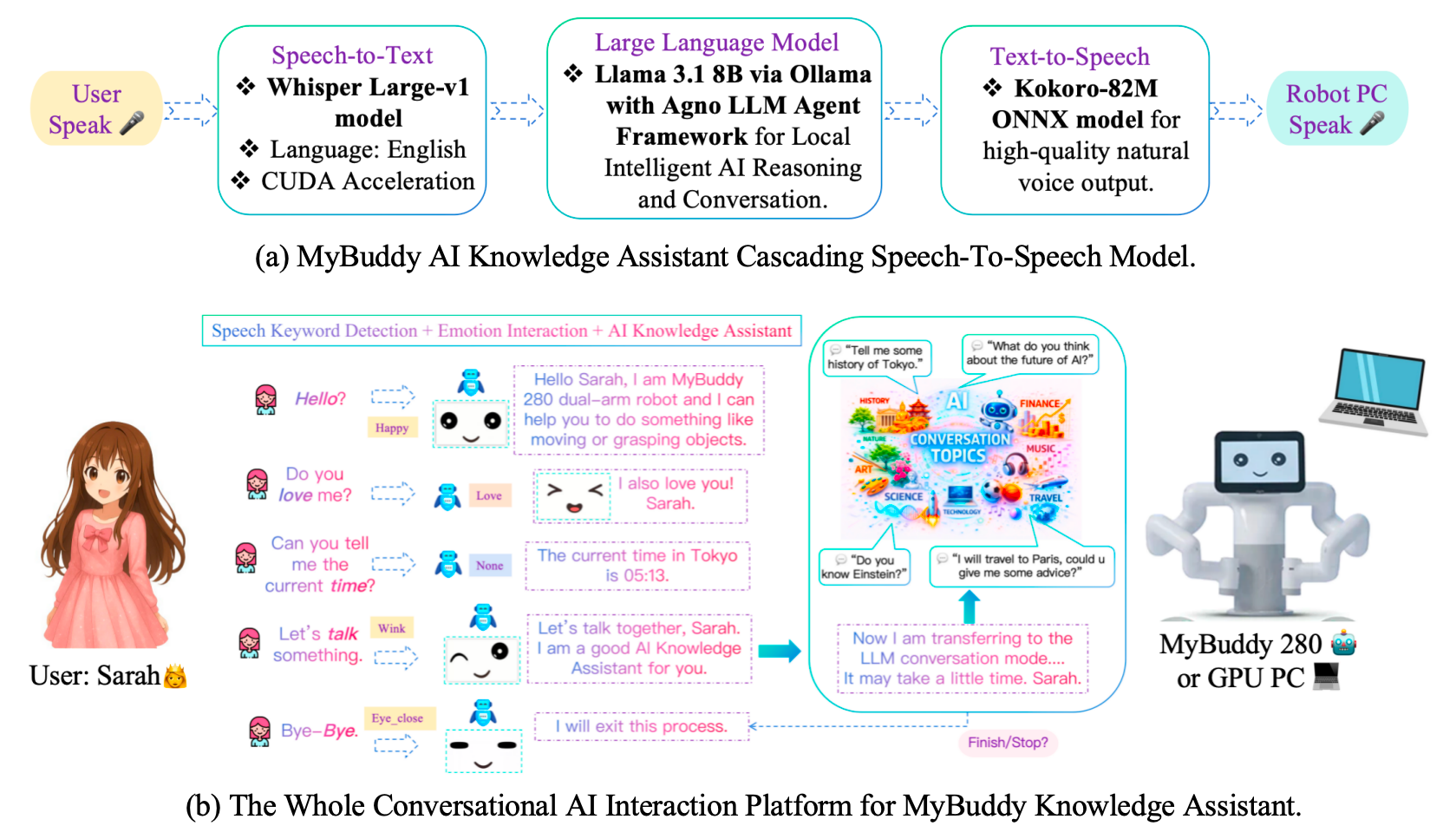}
  \caption{MyBuddy AI Knowledge Assistant Human-Robot Interaction System.}
\end{figure}

Based on the physical dual-arm humanoid robot MyBuddy 280 and the robot-installed USB Microphone, we originally develop the intelligent Conversational AI platform between our MyBuddy 280 robot and human users in Fig. 1. Firstly, we utilize the “Speech Recognition” Python package to make robots recognize what people have spoken, then apply speech keyword detection and emotional interaction techniques to give users diverse corresponding emotion-displayed voice answers according to different speech keywords including “hello” “love” “time” “talk” and “bye”, even establish an advanced LLM-based AI Knowledge Assistant for continuous high-quality Human-Robot real-time speech-to-speech dialogue interaction via intelligent AI reasoning. In detail, our LLM-based AI Knowledge Assistant for MyBuddy 280 is mainly based on the cascading speech-to-speech model architecture as illustrated in Fig. 1 (a), which can outperform normal direct speech-to-speech models via the following key innovations: (1) Better accuracy and faster debugging with each individual component (e.g., ASR, LLM, TTS) optimized for specific tasks. (2) Higher reliability and better AI reasoning capability with the integrated state-of-the-art Large Language Models on information processing. (3) Easier Integration with External Internet Tools (e.g., Google Search) for continuous voice assistant across a truly human-like conversational experience. Powered by the Silero VAD detector and Whisper-Fast transcription model, our MyBuddy AI Knowledge Assistant can successfully recognize the real-time user input speech information and transform it into text for the following LLM-based AI reasoning module, which implements multi-modal reasoning with Llama 3.1 8B model [7] via Ollama based on the Agno LLM agent to generate thoughtful relevant responses. After that, our AI agent can also employ the state-of-the-art Kokoro-82M Text-to-Speech (TTS) model [8] to produce a natural-sounding elegant voice to answer human users with the LLM-generated intelligent text responses. To emphasize, our applied Agno LLM agent even includes extensible tool-calling capabilities of accessing Google Search, Wikipedia, and arXiv for real-time information to generate much perfect AI reasoning answers for wonderful integral conversations.

\section{Experiments}
Transferred to the application experiments, our MyBuddy 280 humanoid robot can not only respond to human users via diverse speech answers and rich facial video emotions (“Happy”, “Love”, “None”, “Wink”, “Eye\_Close”) corresponding to the five defined keywords “hello” “love” “time” “talk” and “bye” within Fig. 1 (b), but also autonomously switch to the intelligent MyBuddy AI Knowledge Assistant mode if the user’s dialogue includes “talk” such as “Let’s talk something”, which can apply the large language model Llama 3.1 8B via Ollama with Agno LLM Agent framework for local natural conversations and intelligent AI reasoning. Obviously in Fig. 1 (b), driven by versatile Large Language Models, human users can discuss with MyBuddy 280 robot via GPU PC about any possible various topics including city weather, current time, Europe travel suggestions, history, scientists, finance, AI, celebrities, art, hot news and nature. In addition, our MyBuddy AI Knowledge Assistant can quickly answer the most appropriate information for users, even provide much more high-quality answers with the integration of Google web search and extensible tool systems (e.g., arXiv, Wikipedia) for seamless voice interactions. 

\begin{figure}
  \centering
  \includegraphics[width=\textwidth]{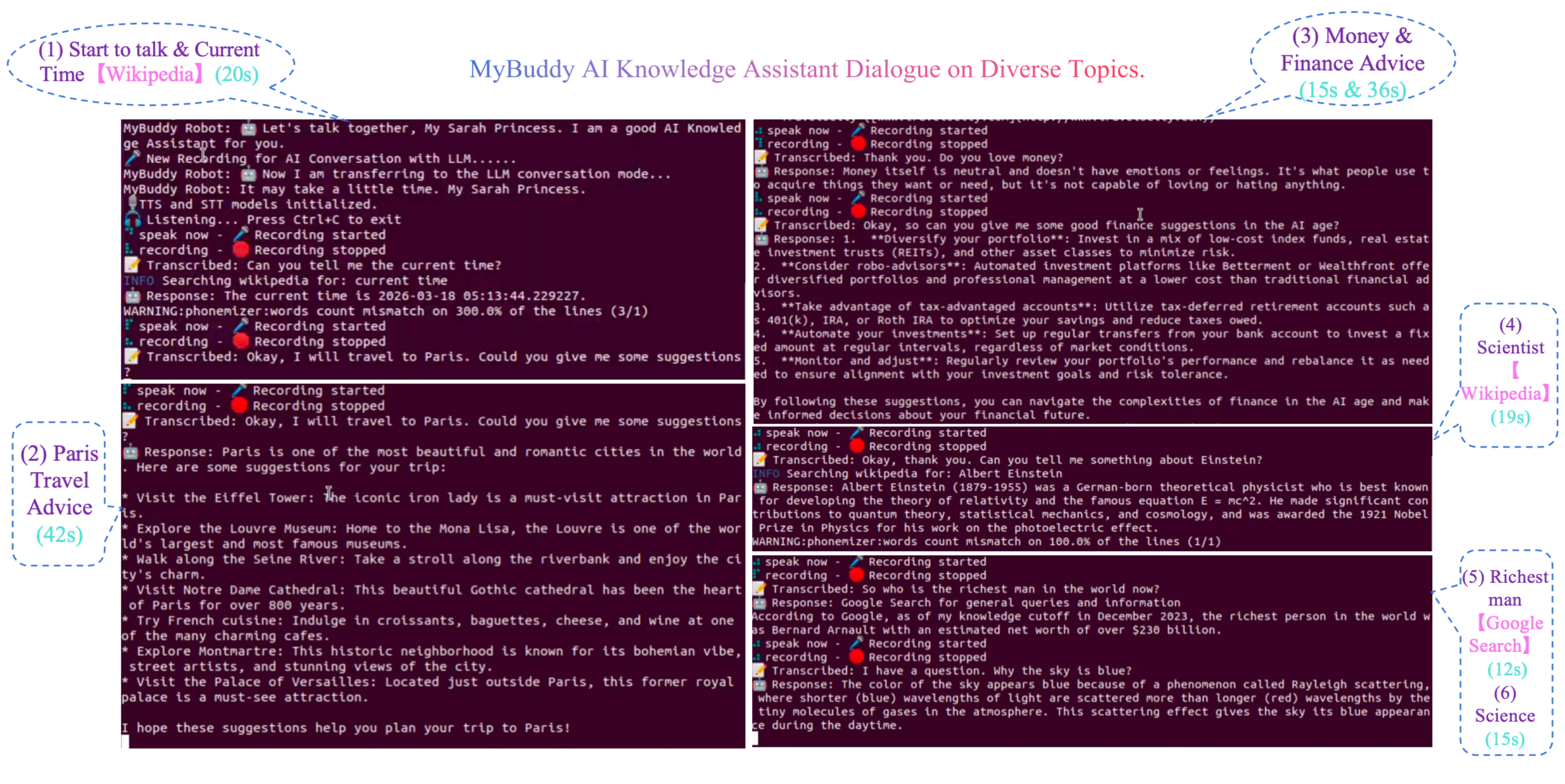}
  \caption{MyBuddy AI Knowledge Assistant dialogue experiments on diverse topics.}
\end{figure}

Furthermore, we intuitively demonstrate the user dialogue process with MyBuddy AI Knowledge Assistant on diverse topics via Fig. 2, even compute the LLM model execution time for each sub-conversation via quantitative evaluation and record the whole conversational AI experimental video as in the link \url{https://drive.google.com/file/d/16nZhDXAB3ObKjobZJLx5VYsCQd9j965U/view?usp=drive_link}. Based on such extensive AI interactive communication experiments, we harvest the following significant discoveries: (1) Our deployed Large Language Model Llama 3.1 8B can generate complete and reasonable responses for any discussion topics including travel, art, history, universe, science, sports, music and any all. (2) For different topics, it may take LLM different execution time to generate the corresponding answers, which exactly depends on the length, difficulty and complexity of the query problem and whether LLMs need to apply Wikipedia or Google for further real-time web search. At first, the user asks “Can you tell me the current time?” and our LLM takes 20s to search the accurate present time with Wikipedia; then later for a little complicated question “I will travel to Paris. Could you give me some suggestions?”, it takes much longer time about 42s to present a comprehensive response with diverse famous attractions in Paris (e.g., Eiffel Tower, The Louvre Museum). Next, our MyBuddy can quickly talk about money and finance advice separately in 15s \& 36s, even utilize the Wikipedia to search related information about the famous scientist Albert Einstein and apply Google Search to find the richest man all around the world in just 12s, which indeed demonstrates the efficiency and effectiveness of Large Language Models within our interactive MyBuddy AI Knowledge Assistant. (3) It’s possible for human users to interact with the MyBuddy 280 at any time as long as they want and the GPU-driven PC computer can assist LLMs to perform much better than CPU devices as illustrated in our provided supplemental experiment demonstration video.

\section{Conclusion and Future Work}
In sum, our innovatively developed real-time cascading speech-to-speech AI Knowledge Assistant, which combines advanced speech recognition, AI reasoning, and natural text-to-speech capabilities, can be easily transferred and applied into any other robotic systems with much more extended speech keywords for intelligent continuous human-like friendly dialogues within diverse life or work interactive scenarios to achieve high-quality knowledge query assistant. For future work, we can integrate the facile real-time Vision-Language Model for more intelligent comprehensive environment analysis into our normal knowledge assistant Conversational AI system, even extend the speech keywords via richer facial emotions within multiple scenarios for natural multi-modal interaction.

\section*{References}

{
\small

[1] Chen H, Wang J, Meng M Q H. Kinova gemini: Interactive robot grasping with visual reasoning and conversational AI[C]//2022 IEEE International Conference on Robotics and Biomimetics (ROBIO). IEEE, 2022: 129-134. 

[2] Neto I, Hu Y, Correia F, et al. "I'm Not Touching You. It's The Robot!": Inclusion Through A Touch-Based Robot Among Mixed-Visual Ability Children[C]//Proceedings of the 2024 ACM/IEEE International Conference on Human-Robot Interaction. 2024: 511-521.

[3] Chen H. Motion Control of Interactive Robotic Arms Based on Mixed Reality Development[C]//Proceedings of the 2023 6th International Conference on Robot Systems and Applications. 2023: 86-93.

[4] Huang A, Ranucci A, Stogsdill A, et al. (gestures vaguely): The effects of robots’ use of abstract pointing gestures in large-scale environments[C]//Proceedings of the 2024 ACM/IEEE International Conference on Human-Robot Interaction. 2024: 293-302. 

[5] Chen H. Robotic manipulation with reinforcement learning, state representation learning, and imitation learning (student abstract)[C]//Proceedings of the aaai conference on artificial intelligence. 2021, 35(18): 15769-15770. 

[6] CHEN H. Robotic Manipulation with Reinforcement Learning, State Representation Learning, and Imitation Learning[J]. 2025.

[7] Grattafiori A, Dubey A, Jauhri A, et al. The llama 3 herd of models[J]. arXiv preprint arXiv:2407.21783, 2024.

[8] Li Y A, Han C, Raghavan V, et al. Styletts 2: Towards human-level text-to-speech through style diffusion and adversarial training with large speech language models[J]. Advances in neural information processing systems, 2023, 36: 19594-19621.

%%%%%%%%%%%%%%%%%%%%%%%%%%%%%%%%%%%%%%%%%%%%%%%%%%%%%%%%%%%%

\appendix

%%%%%%%%%%%%%%%%%%%%%%%%%%%%%%%%%%%%%%%%%%%%%%%%%%%%%%%%%%%%

%\newpage
%\input{checklist.tex}

\end{document}